%% file: Incremental_learning_ICRA2020.tex
\documentclass[letterpaper, 10 pt, conference]{ieeeconf}  

\IEEEoverridecommandlockouts                              
\usepackage{graphicx} 
\usepackage{epsfig} 
\usepackage{mathptmx} 
\usepackage{times} 
\usepackage{xcolor}
\usepackage{subfigure}
\usepackage{sidecap}
\usepackage{floatrow}
\usepackage{subfigure}
\usepackage{amsfonts}
\usepackage{amsmath,amssymb,balance}
\usepackage[compress]{cite}
\usepackage{bm}
\usepackage[linesnumbered,ruled,vlined]{algorithm2e}

\newcommand{\ghcomment}[1]{\textbf{\textcolor{orange}{G.H. #1}}}
\newcommand{\njcomment}[1]{\textbf{\textcolor{blue}{N.J. #1}}}
\newcommand{\mecomment}[1]{\textbf{\textcolor{magenta}{M.E. #1}}}

\newcommand{\XX}[1]{\textcolor{red}{#1}}
 \renewcommand{\ghcomment}[1]{}
  \renewcommand{\njcomment}[1]{}
  \renewcommand{\XX}[1]{}
  \renewcommand{\mecomment}[1]{}

\usepackage[normalem]{ulem}

\newcommand{\topic}[1]{\textbf{\textcolor{magenta}{#1}}}
\renewcommand{\topic}[1]{}

\title{\LARGE \bf Incremental Learning of Motion Primitives for\\%
Pedestrian Trajectory Prediction at Intersections}

\author{Golnaz Habibi$^{1}$ and Nikita Jaipuria$^{2}$ and Jonathan P. How$^{1}$
\thanks{*This work was supported by Ford Motor Company}
\thanks{$^{1}$Jonathan P. How and Golnaz Habibi are with the Department of Aeronautics and Astronautics,
        MIT. 
{\tt\small golnaz,jhow@mit.edu}}%
\thanks{$^{2}$ Nikita Jaipuria is with Ford Motor Company and contributed to this work during her time in MIT. 
{\tt\small niksjaipuria@gmail.com}}%
}

\begin{document}
\maketitle



\thispagestyle{empty}
\pagestyle{empty}
\begin{abstract}
This paper presents a novel incremental learning algorithm for pedestrian motion prediction, with the ability to improve the learned model over time when data is incrementally available. In this setup, trajectories are modeled as simple segments called motion primitives. Transitions between motion primitives are modeled as Gaussian Processes. When new data is available, the motion primitives learned from the new data are compared with the previous ones by measuring the inner product of the motion primitive vectors. Similar motion primitives and transitions are fused and novel motion primitives are added to capture newly observed behaviors. The proposed approach is tested and compared with other baselines in intersection scenarios where the data is incrementally available either from a single intersection or from multiple intersections with different geometries. In both cases, our method incrementally learns motion patterns and outperforms the offline learning approach in terms of prediction errors. The results also show that the model size in our algorithm grows at a much lower rate than standard incremental learning, where newly learned motion primitives and transitions are simply accumulated over time.
\end{abstract}

\input{introduction}

\input{related_work}
\input{background}
\input{algorithm}
\input{result}
\input{conclusion}

\balance
\bibliographystyle{IEEEtran}

\end{document}

%% file: introduction.tex
\section{INTRODUCTION}
\topic{challenges in pedestrian prediction}
Safe navigation of self-driving vehicles and robots in urban environments requires the ability to interact with, and accurately predict the motion of, other moving agents, including cars, cyclists and pedestrians. Pedestrian motion prediction is challenging as compared to that of cars and cyclists, as the ``rules'' as less clear and more frequently violated. Pedestrians can also abruptly change their direction of motion due to environmental contexts; and modeling all such contexts is often impractical. The complexity is increased further in urban environments (e.g., intersections), where additional context, such as traffic lights, stop signs and environment geometry, such as location of sidewalks and crosswalks, also influences pedestrian movement~\cite{2018arXiv180609444J}.

\topic{challenges in pedestrian prediction}
Thanks to larger, annotated datasets and faster computers, data-driven machine learning techniques have become popular tools for learning motion behaviors in urban areas~\cite{ridel2018literature}. However, most of these techniques are limited to the classical batch setting where a dataset comprising of a wide variety of behaviors is used to train models offline, while inference, i.e.\ behavior prediction, is run online~\cite{losing2018incremental}. Such a setting works well if the training dataset is representative of all possible behaviours one would expect to encounter during inference. But the large diversity in human behaviors makes offline learning impractical as autonomous robots and vehicles may encounter different behaviors as they explore new environments. Offline training in such environments does not have the option of learning new motion behaviours \emph{incrementally} and hence, limits the adaptation of the pre-trained model to new behaviors.

Incremental learning refers to learning from streaming data, that is available incrementally over time. Such a setting offers systems, with limited memory or computation power, the possibility of processing big data gradually in scenarios where offline learning is challenging and impractical. Prior works in incremental learning of motion behaviors either need predefined motion patterns~\cite{kulic2008incremental} or require complete trajectories, \emph{i.e.}, trajectories including the pedestrian's point of entry and exit from an intersection, for training and inferring pedestrian intent (goal)~\cite{vasquez2009incremental}. Collecting such datasets is impractical in busy intersections where often a part of the pedestrian trajectory is occluded by cars, buildings and/or other pedestrians.
 
This paper presents the Similarity-based Incremental Learning algorithm (SILA) for building models for accurate pedestrian motion prediction in urban intersections. Our prediction models comprise of \emph{motion primitives} and pair-wise \emph{transitions} between them. 
To learn motion primitives that can be transferred across environments (i.e.\ intersections with different geometries), first, the observed trajectories are projected from the original frame into a common frame~\cite{2018arXiv180609444J}. Then, motion primitives and the pair-wise transitions between them are learned using a sparse coding technique~\cite{chen2016augmented}. To reduce computational complexity, the transitions are modeled as Sparse Gaussian Processes (GPs) based on ``pseudo inputs"~\cite{snelson2006sparse}.

\topic{how they use it before}


Normalized inner product of motion primitives from each model is applied to measure the pair-wise similarity between pre-trained and new motion primitives and a fusion strategy based on \emph{similarity graph} is proposed. 
The following summarizes the main contributions:
\begin{itemize}
\item Present SILA to incrementally update the prediction model while simultaneously \emph{transferring} knowledge across intersections with different geometries.
\item 
Use normalized inner product techniques to measure similarity between motion primitives learned across intersections with different geometries to constrain model size growth.
\item Show that SILA has significantly slower growth in model size compared to standard incremental learning.
\item Show that significantly lower learning time enables SILA for online learning of motion behaviors. 
\end{itemize}

%% file: related_work.tex
\section{RELATED WORK}
 This section briefly reviews the previous work in incremental learning applied to motion prediction. 
Ref.~\cite{vasquez2009incremental} introduced a Growing Hidden Markov Model (GHMM) based incremental learning algorithm for pedestrian motion prediction. Their method relies on estimating pedestrian intent for trajectory prediction and hence, requires full pedestrian trajectories, from intersection entry till exit, to train on. In contrast, SILA can learn from incomplete or partially occluded trajectories~\cite{chen2016augmented}. Ref.~\cite{kulic2008incremental} incrementally learns the relationship between motion primitives for body movements using HMMs. However, the set of motion primitives itself is pre-defined. Such an approach does not directly apply to the task of pedestrian trajectory prediction, where the wide variety of motion primitives cannot be pre-defined and must instead be learned from data. Thus, SILA incrementally adds novel primitives to the model, while simultaneously updating existing primitives. Ref.~\cite{ferguson2015real} does online learning of motion patterns by detecting changing intent using Gaussian Processes. In their work, when a new motion behavior is observed, it is compared with the pre-trained model and if different, it is considered as a new behavior/change in intent. SILA, on the other hand, not only detects new behaviors and adds those to the model, it also fuses similar behaviors to enrich the learning model. Ref.~\cite{chen2016augmented} provides a compact representation of motion behaviors in the form of motion primitives or dictionary atoms which cluster trajectories into local segments. The motion patterns of these clusters and their transitions are modeled as Gaussian Processes to predict pedestrian motions. This paper leverages the idea of \cite{chen2016augmented} to compactly represent pedestrian motion in terms of motion primitives and their transitions. 




%% file: background.tex
\section{BACKGROUND AND NOTATIONS}
\label{sec:background}
\begin{figure*}[t]
\centering
\subfigure[]{\label{fig:dictionary1}\includegraphics[width = 0.25\linewidth]{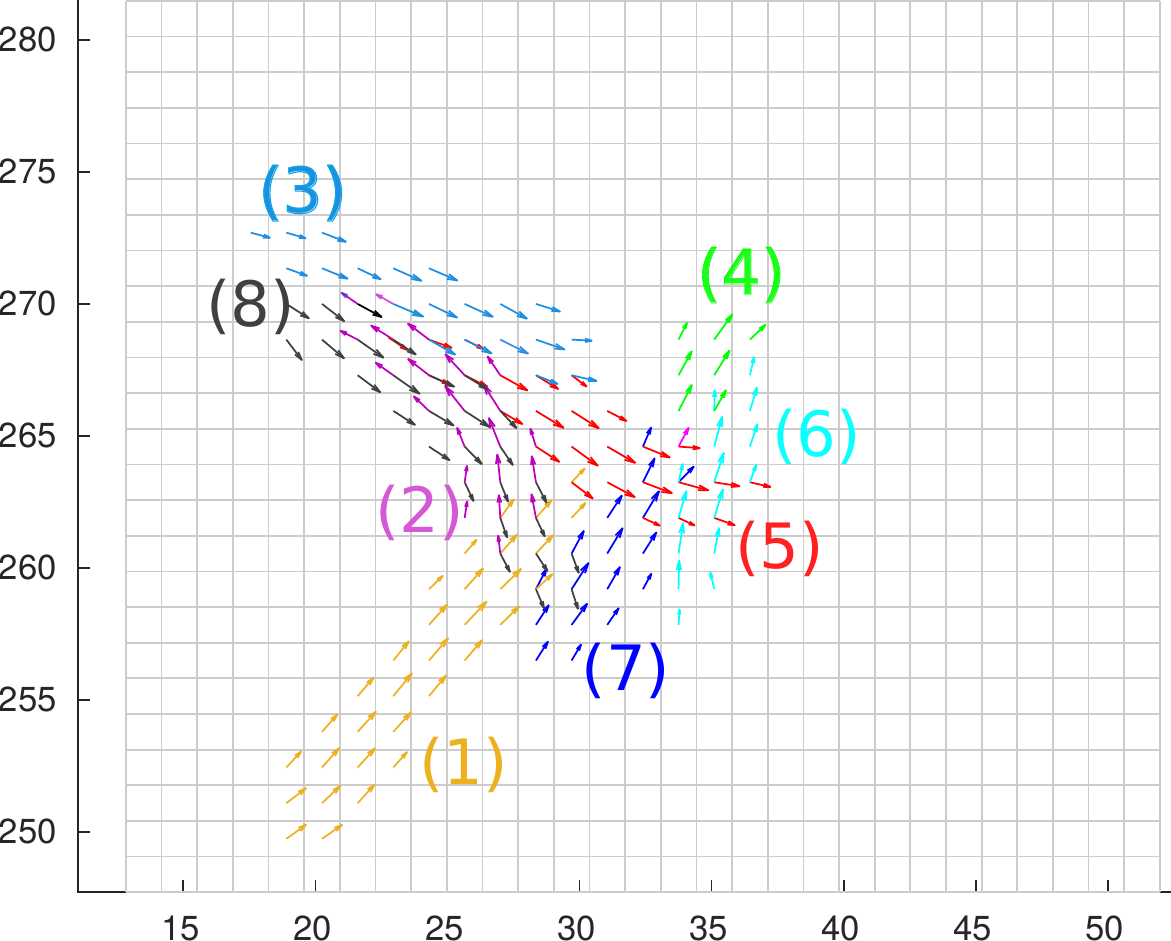}}
\quad \subfigure[]{\label{fig:trajsegment}\includegraphics[width = 0.25\linewidth]{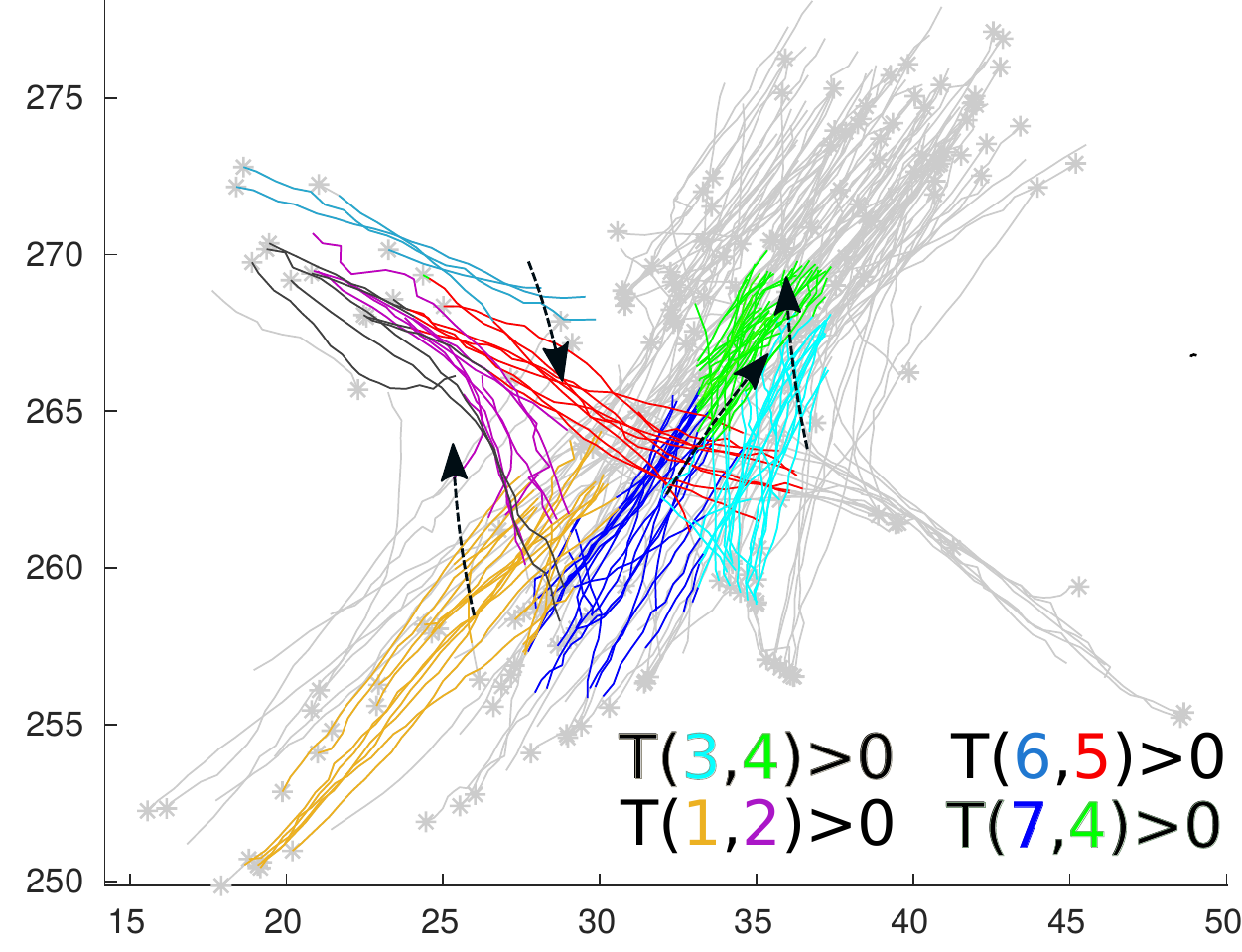}}
\quad \subfigure[]{\label{fig:dual}\includegraphics[width = 0.16\linewidth]{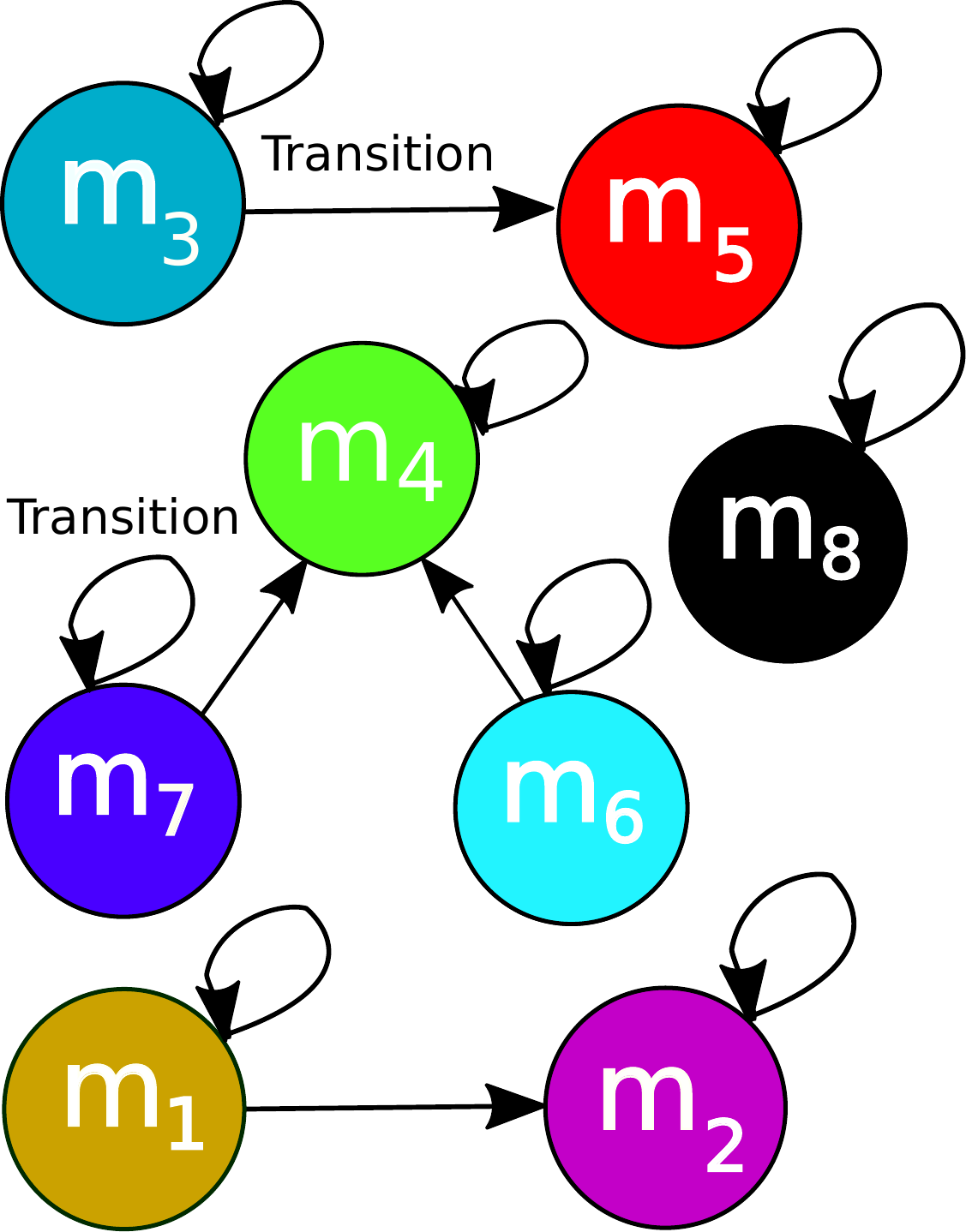}}
\quad \subfigure[]{\label{fig:split}\includegraphics[width = 0.11\linewidth]{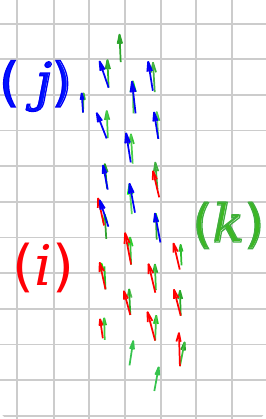}}
\quad \subfigure[]{\label{fig:inout}\includegraphics[width = 0.11\linewidth]{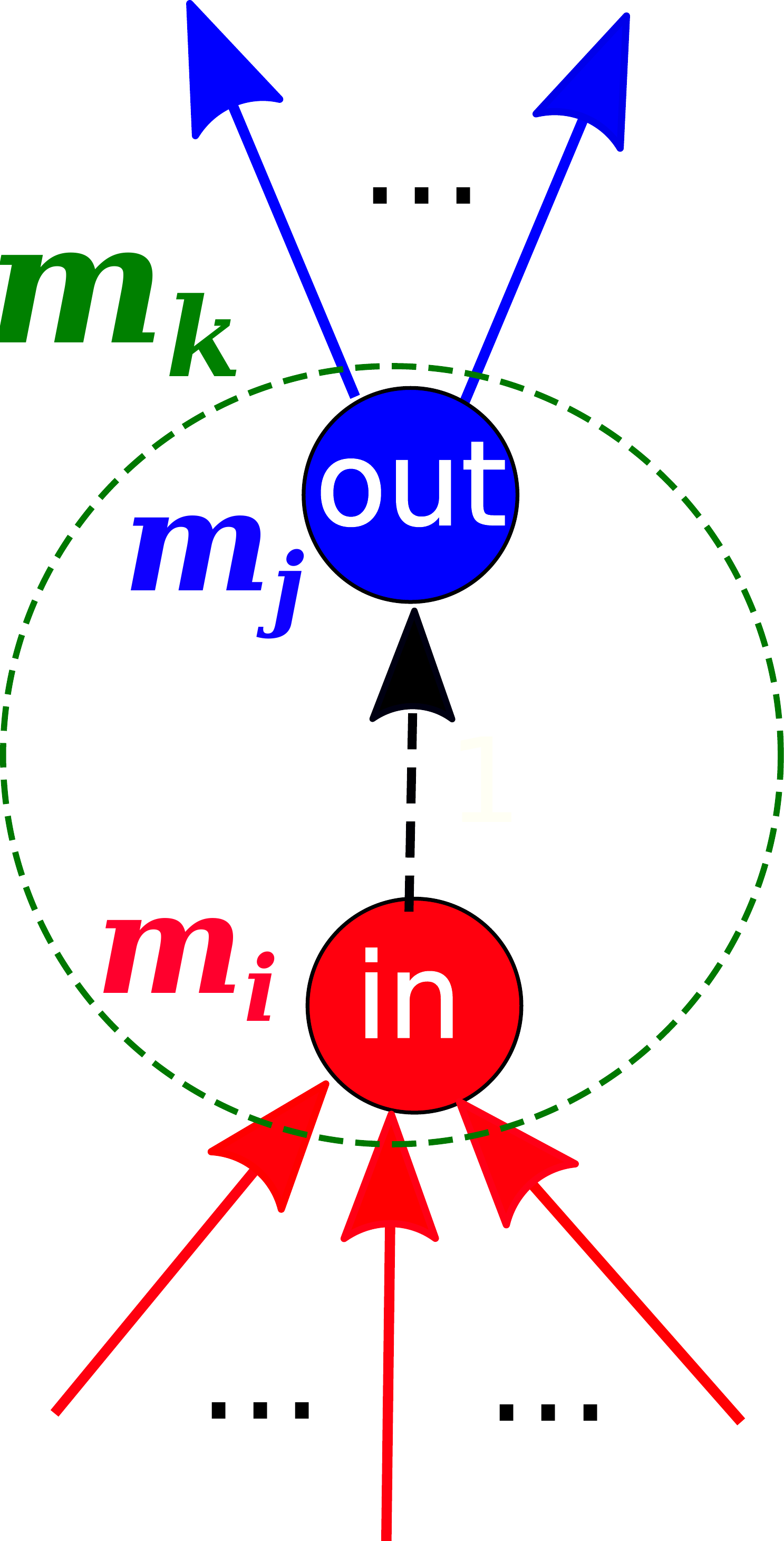}}
\vspace*{-.15in}
\caption{ \label{fig:dictionary} \textbf{(a)} Motion primitives are learned in a grid world using ASNSC~\cite{chen2016augmented}. Each color represents a single motion primitive $\mathbf{m}_i$ learned from pedestrian trajectories; \textbf{(b)} Segmentation of training trajectories (gray) into clusters,  each cluster is best explained by the motion primitive of the same color in (a), the black arrows show the transition between motion primitives, for cases in which the corresponding element in the transition matrix $\mathbf{T}$ is non-zero; \textbf{(c)} motion primitives and their transitions in (b) are modeled as a directed graph, called \emph{motion primitive graph} where nodes represent motion primitives and edges represent transitions, corresponding to only the 8 motion primitives shown in (a), not the entire learned set of 29 motion primitives. \textbf{(d)} Case where two motion primitives, $\mathbf{m}_i$ and $\mathbf{m}_j$, are similar to another primitive $\mathbf{m}_k$ (refer Case 1 in Section III-D under connected components with two edges). Here, $\mathbf{m}_i$ and $\mathbf{m}_j$ together provide a richer representation of motion behaviors. Thus, $\mathbf{m}_k$ is replaced by $\mathbf{m}_i$ and $\mathbf{m}_j$, while all of $\mathbf{m}_k$'s relationships to other nodes are modified as shown in the following figure. \textbf{(e)} replacing $\mathbf{m}_k$ with $\mathbf{m}_i$ and $\mathbf{m}_j$, as explained in (d), 
all the edges entering $\mathbf{m}_{k}$ now enter $\mathbf{m}_{i}$; and all edges exiting from $\mathbf{m}_{k}$ now exit from $\mathbf{m}_{j}$, which requires re-indexing as follows: $\mathbf{m}_{k,in} = \mathbf{m}_{i,in}$ and $\mathbf{m}_{k,out} = \mathbf{m}_{j,out}$ (figures are seen better in colors).}
\end{figure*}

As proposed in~\cite{chen2016augmented}, trajectories are mapped into a grid world with $N= r \times c$ cells, where $r$ and $c$ are the number of rows and columns respectively. Let the training dataset consist of $p$ trajectories. The $i$-th trajectory can be represented as a column vector $\mathbf{tr}_{i} \in \mathbb{R}^{N}$ such that the $k$-th element of $\mathbf{tr}_i$ is the normalized velocity vector of the $i$-th trajectory in the $k$-th grid cell. Given this vectorized representation of training trajectories, a set of $L$ motion primitives (dictionary atoms), $\mathbf{D} = \{\mathbf{m}_{1}, \ldots, \mathbf{m}_L\}$, are learned using sparse coding~\cite{chen2016augmented}. Each color in Fig.~\ref{fig:dictionary1} represents a single motion primitive $\mathbf{m}_{i}$, learned from the trajectories shown in gray in Fig.~\ref{fig:trajsegment}. Each motion primitive $\mathbf{m}_{i}$ is represented as a set of normalized cell-wise velocities $\{\mathbf{v}^k_i\}$ . Here, $\mathbf{v}^k_i$ is the velocity of $\mathbf{m}_{i}$ in the k-th grid cell (in $N=25 \times 29$ grid cells in Fig.~\ref{fig:dictionary1}). 

Dictionary atoms ($\mathbf{D}$) are used to segment the training trajectories into clusters (Fig.~\ref{fig:trajsegment}). Each color in Fig.~\ref{fig:trajsegment} is one such cluster, best explained by the motion primitive in Fig.~\ref{fig:dictionary1} in the same color. These clusters are used to create the transition matrix $\mathbf{T}_{L \times L}$, where $T(i,j)$ denotes the number of training trajectories exhibiting a transition from $\mathbf{m}_i$ to $\mathbf{m}_j$. Following Ref.~\cite{chen2016augmented}, each of these transitions is modeled as a two-dimensional GP flow field~\cite{joseph2011bayesian,aoude2013probabilistically, snelson2006sparse}. Thus, $\mathbf{T}$ is used to create the set of transitions $\mathbf{R}$, which has two components: $\mathbf{R}_{edges} =\{(i, j)|\mathbf{T}(i,j)\neq \emptyset\}$ and $\mathbf{R}_{GP}$. $\mathbf{R}_{GP}$ consists of two independent GPs for each tuple in $\mathbf{R}_{edges}$, $(GP_x,GP_y)$, 
that learn a mapping from the two-dimensional position features $(x,y)$ to velocities $v_x$ and $v_y$ respectively~\cite{Chen2017}. These GPs are also referred to as `motion patterns'.
We use a sparse GP regression model for learning motion patterns~\cite{snelson2006sparse}. Such a model uses a fixed number of pseudo-inputs, learned using a gradient-based optimization, as a sparse representation of the full GP. The hyperparameters of the covariance function are also estimated using a joint optimization technique, thus doing away with the need to manually tune hyperparameters, as in previous GP models~\cite{csato2002sparse}.

Assume $M(n-1)$ is the trained model at the $(n-1)$-th training episode and comprises of the set of motion primitives $\mathbf{D}=\{\mathbf{m}_{1},\cdots,\mathbf{m}_{N_1}\}$ and the set of transitions $\mathbf{R}$, i.e.\ $\mathbf{R}_{edges}$ and $\mathbf{R}_{GP}$. 
In the $n$-th training episode, a new batch of data is used to learn $M'(n)$, comprising of $\mathbf{D}'=\{\mathbf{m'}_1,\cdots,\mathbf{m'}_{N'_1}\}$ motion primitives and $\mathbf{R}'$ transitions with $\mathbf{R}'_{edges}$ and $\mathbf{R}'_{GP}$. The following section explains SILA algorithm for updating $M(n-1)$ to $M(n)$ by computing the similarity between the incremental model $M'(n)$ and $M(n-1)$; and fuse them based on a similarity graph.

%% file: algorithm.tex
\section{SIMILARITY-BASED INCREMENTAL LEARNING ALGORITHM (SILA)}
This section describes an algorithm for motion prediction (Algorithm  \ref{alg:sila}) that incrementally updates the model as more data is available, without the need to re-learn the full model from scratch. We demonstrate its capability on the specific use-case of pedestrian trajectory prediction in urban intersections. SILA has five main steps in each training episode: (1) normalize trajectory data using~\cite{jaipuria2018learning}; (2) learn $M'(n)$ from new data using ~\cite{chen2016augmented}; (3) augment pre-trained model $M(n-1)$ with $M'(n)$ to get $\hat{M}(n)$ and re-index the motion primitives and their transitions (line \ref{line:initialize}); (4) measure similarity between motion primitives in $M'(n)$ and $M(n-1)$ to create a similarity graph (line \ref{line:similarity}); (5) fuse $M(n-1)$ with $M'(n)$ based on the similarity graph to get $M(n)$ (line \ref{line:casesstart}-\ref{line:fuseend}). The rest of this section describes each of these steps in detail. Steps 1 and 2 are prepossessing stages from existing works. The main contribution of this paper are steps 3-5.
\subsection{Normalize Trajectory Data\label{sec:normalize}}
We use the Augmented Semi Non-negative Sparse Coding (ASNSC) algorithm~\cite{chen2016augmented} to learn trajectory prediction models as a set of motion primitives and the pair-wise transitions between them, using spatial position and velocity features. Thus, to learn from data collected in different intersections (each with its own frame of reference), it becomes imperative to learn motion primitives and transitions from features that are transferable across environments. This issue is addressed by the use of a common frame~\cite{jaipuria2018learning}, with its origin at the intersection corner of interest and axes aligned with the intersecting curbsides. In this common frame, each data point is represented by its \emph{contravariant distance}~\cite{bowen2008introduction} from the curbsides (see Fig.~\ref{fig:intersections}). Furthermore, the transformed trajectories are normalized with respect to the sidewalk width to account for scale differences across intersections.

\subsection{Learn Motion Primitives and Transitions} 
In every training episode, ASNSC learns motion primitives and transitions from the new batch of data available, which are then modelled as a \emph{motion primitive graph}~\cite{kulic2012incremental}. In this graph (Fig.~\ref{fig:dual}), nodes represent a motion primitive and edges represent the order of observing primitives, i.e.\ transitions.

\subsection{Initialize Indexes of Motion Primitives and Transitions}
SILA's ultimate goal is to efficiently merge the pre-trained model $M(n-1)$ with the new model $M'(n)$ to get a single model $M(n)$ that is best representative of all motion behaviours seen so far. The first step towards this goal is to re-index the motion primitives and transitions from $M(n-1)$ and $M'(n)$ to avoid repetition of indices. Thus, initially the accumulated model $\hat{M}(n)$ simply combines $M(n-1)$ with $M'(n)$, such that $\mathbf{\hat{D}} = \{\mathbf{D},\mathbf{D}'\}$ is of size $N_{1}+N'_{1}$. Similarly, the set of accumulated transitions $\mathbf{\hat{R}} = \{\mathbf{R},\mathbf{R}'\} $ is of size $N_{2}+N'_{2}$
(Alg. ~\ref{alg:sila}, line~\ref{line:initialize}) 
 
Each node is represented by two \emph{virtual nodes}, denoted by $in$ and $out$, and a \emph{virtual transition} (edge): $(in, out)$. Given this \emph{virtual} representation of each node, we need to re-index edges such that all edges entering $\mathbf{m}_{k}$ now enter $\mathbf{m}_{k,in}$; and all edges exiting $\mathbf{m}_{k}$ now exit $\mathbf{m}_{k,out}$. This implies that for every transition edge $(i,j)$, (\emph{i.e.} edges exiting from $\mathbf{m}_i$ and entering $\mathbf{m}_j$, in  $\mathbf{\hat{R}}_{edges}$ re-indexing is done as follows: $(i,j) \mapsto (\mathbf{m}_{i,out}, \mathbf{m}_{j,in})$. In the beginning of each learning episode, indices $in$ and $out$ for all nodes are initialized equivalently. For instance, $\mathbf{m}_{k,in} = \mathbf{m}_{k,out} = k$ for the $k$-th motion primitive in accumulated motion primitive set.
Fig.~\ref{fig:similar_graph} shows the accumulated motion primitive graph of $M'(n)$ and $M(n-1)$, i.e., $\hat{M}(n)$, post re-indexing (\emph{i.e.}, each index is unique and not repeated). The numbers in circles denote the motion primitive index. For clarity, nodes are colored based on the model they come from, and transitions are shown as black arrows.

A special set, called the \emph{fusion set}, is also defined and initialized as an empty set in this step. Eventually, it contains a list of tuples of nodes that need to be fused. For \emph{e.g.}\ the tuple $(\mathbf{m}_i,\mathbf{m}_j)$ implies $\mathbf{m}_i$ and $\mathbf{m}_j$ need to be fused in the model fusion step.
\subsection{Similarity between Motion Primitives and Re-indexing of Motion Primitives and Transitions} 
The next step is to compute the similarity between motion primitives in $M(n-1)$ and $M'(n)$. 
Motion primitives are representative of human motion behaviors in terms of both heading and region of occurrence in the grid-based world. Thus, computing the similarity between such motion primitives should reflect both the difference in heading and the overlap of regions of occurrence.
In this work, similarity between motion primitives is computed as the cosine of the angle between the two motion primitive vectors, which is equivalent to their normalized inner product:
\begin{equation}
\label{eq:mutual2}
S(\mathbf{m}_{i},\mathbf{m'}_{j}) = \frac{\langle\mathbf{m}_{i},\mathbf{m'}_{j}\rangle}
{|\mathbf{m}_{i}||\mathbf{m'}_{j}|}
\end{equation}
where, $\langle\cdot,\cdot\rangle$ denotes inner product, $\mathbf{m}_i$ is the $i$-th motion primitive in $M(n-1)$, $\mathbf{m'}_j$ is $j$-th motion primitive in $M'(n)$.
Here, $S(\mathbf{m}_{i},\mathbf{m'}_{j})$ accounts (in a grid environment) for the number of overlapping cells between two motion primitives as well as the difference between the angle of their direction in each cell.

Given Eq.~\ref{eq:mutual2}, pairs of nodes $M(n-1)$ and $M'(n)$ with similarities greater than a pre-defined threshold, $t_s$, are considered as \emph{matched nodes}.
Fig.~\ref{fig:cc} shows the \emph{similarity graph}~\cite{zager2008graph} for $M(n-1)$ and $M'(n)$, denoted by $G_{s}$, where nodes represent motion primitives, and edges represent the similarity between motion primitives. An edge exists only if the similarity is greater than $t_s$ and is assigned a weight that is equal to the similarity value:
\begin{equation}
e(i,j') = \{(i,j') | S(\mathbf{m}_{i},\mathbf{m'}_j)\geq t_s\}
,\omega(i,j') = S(\mathbf{m}_{i},\mathbf{m'}_j)
\end{equation}
Each set of matching nodes creates a connected component in $G_s$~\cite{Hopcroft}. Fig.~\ref{fig:cc} shows $G_s$ with its connected components encircled in red. $G_s$ by definition does not include unmatched nodes, which implies that the unmatched nodes are neither re-indexed nor merged. The rest of this section presents strategies for re-indexing nodes based on the topology of the connected component.
\subsubsection{Connected components with one edge} 
\begin{figure}[]
\centering
\subfigure[]{\label{fig:similar_graph}\includegraphics[width = 0.35\textwidth]{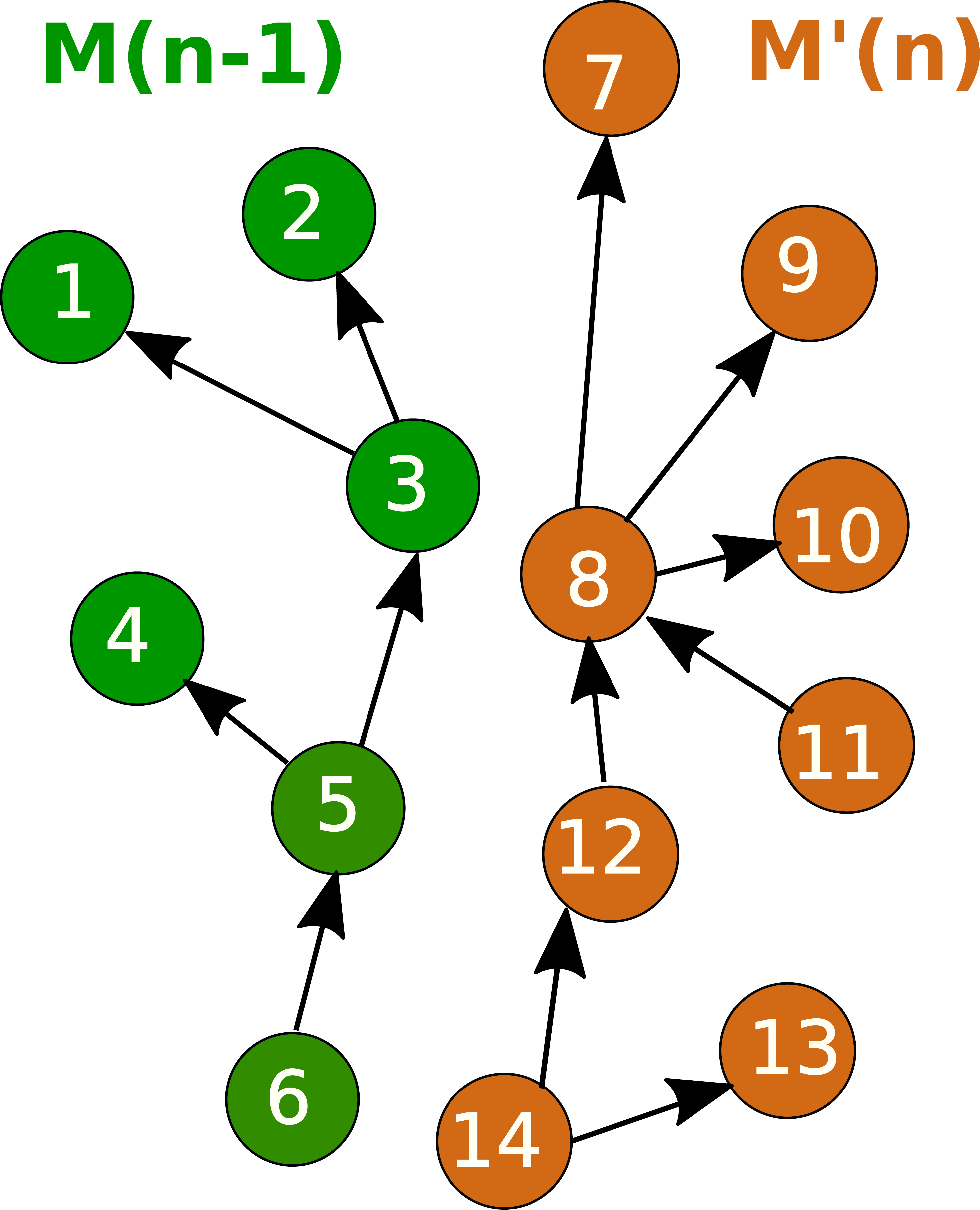}}
\subfigure[]{\label{fig:cc}\includegraphics[width = 0.27\textwidth]{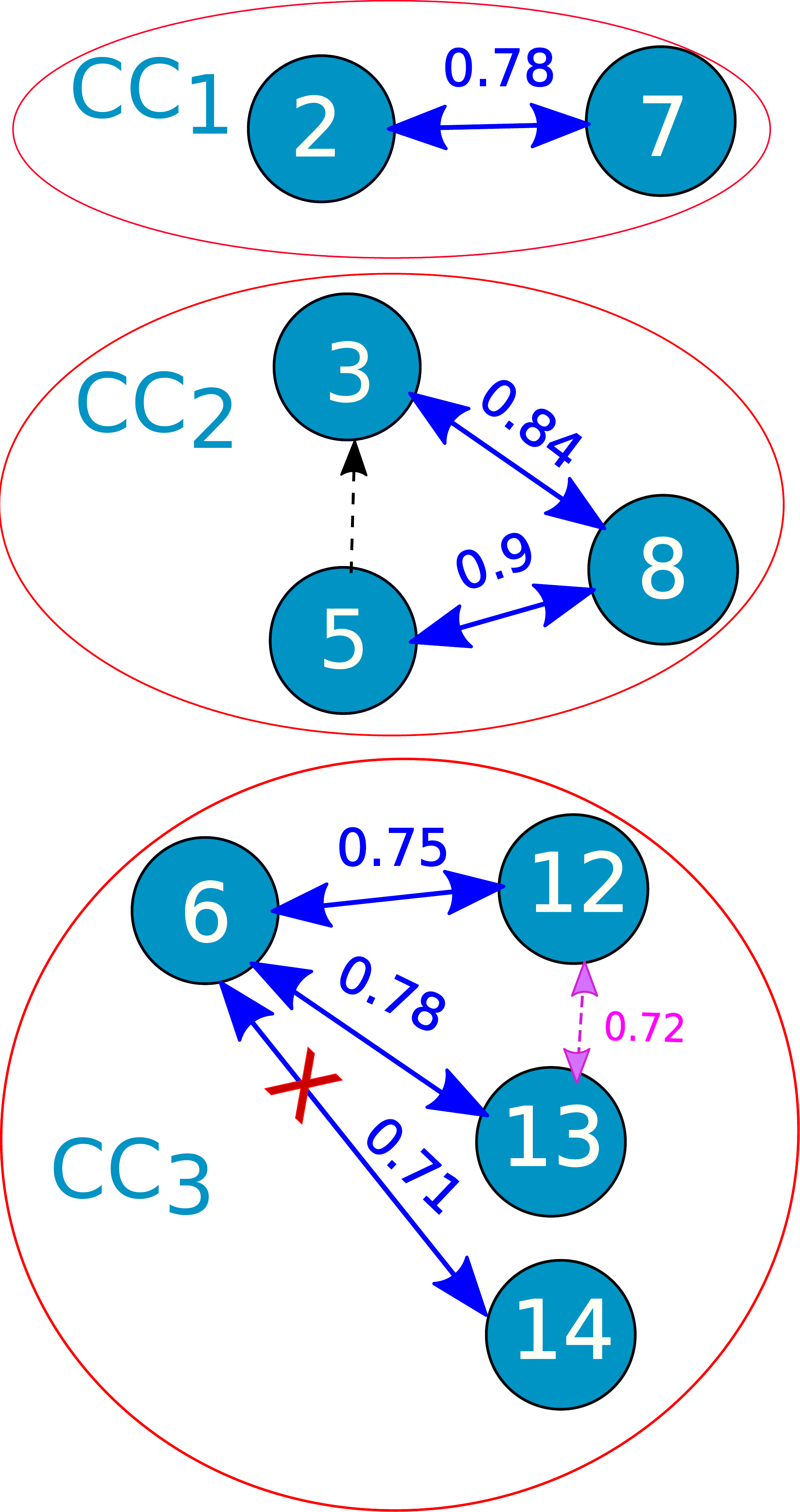}}
\quad\subfigure[]{\label{fig:graphunion}\includegraphics[width = 0.27\textwidth]{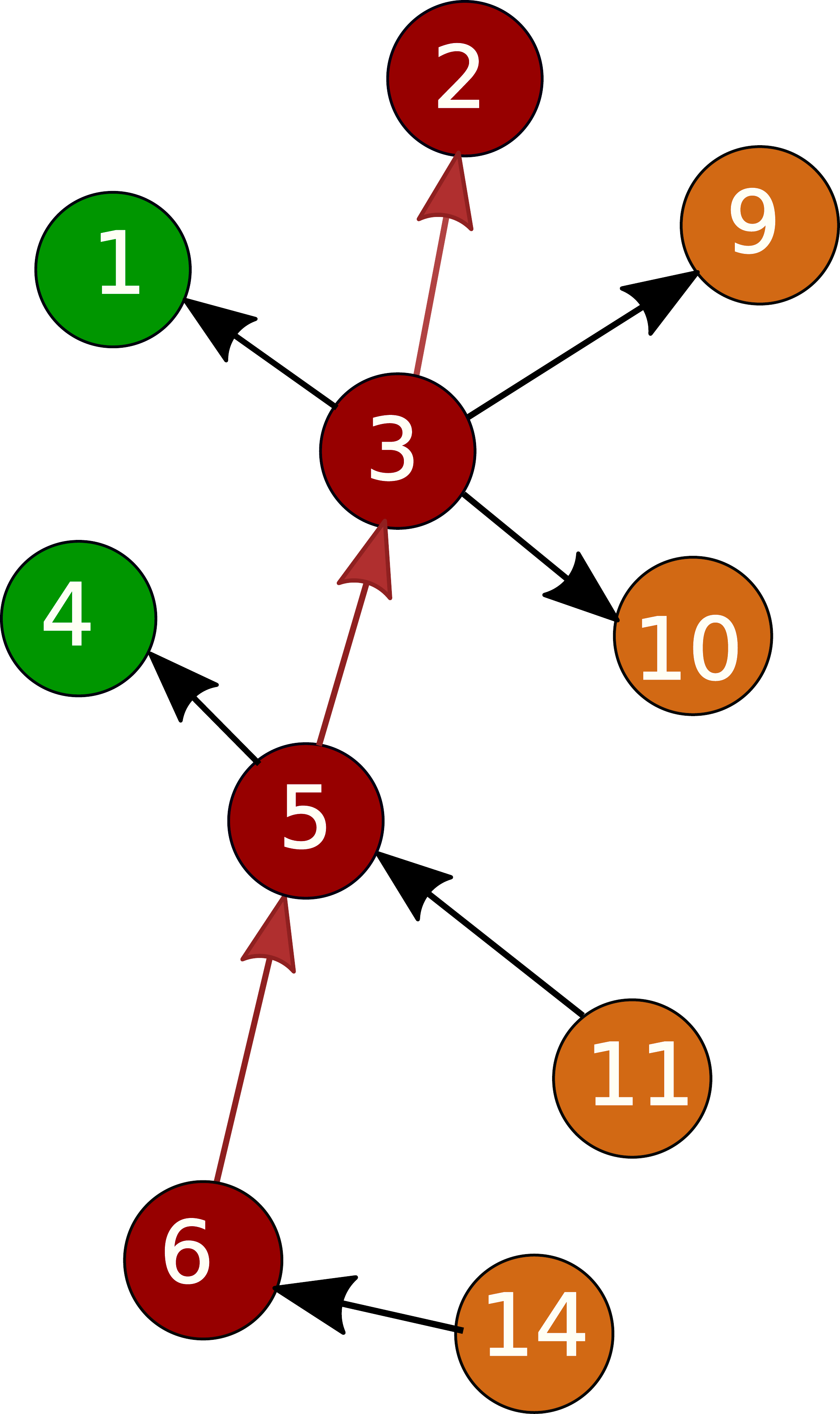}}
\vspace*{-.15in}
\caption{\label{fig:primitivegraph} (a) motion primitive graph of accumulated model $\hat{M(n)}$, pre-trained $M(n-1)$ in green and new model $M'(n)$ is in brown (self transitions are omitted for brevity), black arrows represent the directed transitions. Nodes from both models are re-indexed to have a unique id,(b)similarity graph $G_s$ with threshold of $t_s =0.7$ ,blue two sided arrows are matching relation and nodes are colored in blue regardless of their original model, (note: magenta arrow shows the intra-similarity between two nodes from same model which is considered in case 2 of the algorithm) $G_s$ has three connected components: $CC_1$, $CC_2$, and $CC_3$, in $CC_3$ the edge with least similarity is relaxed (crossed).(c) final motion primitive graph corresponding to updated model $M(n)$ which is created after re-indexing the matched nodes based on the topology of connected components in (b) and applying it in (a) nodes with no fusion are colored by their original model color and fused nodes and edges are colored in dark red. The nodes is finally re-indexed to $1:10$ which is not shown in this figure to help the reader understand all steps. The fusion set in this example is $\{(2,7), (6,12,13)\}$}
\end{figure}

has two nodes $\mathbf{m}_i$ and $\mathbf{m}_j$ (see Fig.~\ref{fig:cc}-$CC_{1}$). This case is also called `one-to-one matching'. To re-index, $in$ and $out$ indices of both nodes are collapsed to the same index, as follows: $\mathbf{m}_{i,in} = \mathbf{m}_{i,out}= \mathbf{m}_{j,in} = \mathbf{m}_{j,out} = min(\mathbf{m}_{i,in},\mathbf{m}_{j.in})$.

\subsubsection{Connected components with two edges}
it happens when two nodes from one model, $\mathbf{m}_i$ and $\mathbf{m}_j$, are matched with a single node $\mathbf{m}_k$ from another model (see Fig.~\ref{fig:cc}-$CC_2$). The re-indexing strategy now depends on the relationship between $\mathbf{m}_i$ and $\mathbf{m}_j$ in their original motion primitive graph, these cases are considered in this following order (Alg. \ref{alg:sila}, lines \ref{line:casesstart}-\ref{line:casesend}):
\begin{itemize}
\item \textbf{Case 1: there exists a transition from $\mathbf{m}_i$ to $\mathbf{m}_j$.} This implies that $\mathbf{m}_i$ and $\mathbf{m}_j$ together have a richer and more primitive representation of motion behaviours as compared to $\mathbf{m}_k$. Thus, $\mathbf{m}_k$ is replaced with $\mathbf{m}_i$ and $\mathbf{m}_j$ and its indices are re-indexed as $\mathbf{m}_{k,in}=\mathbf{m}_{i,in}$ and $\mathbf{m}_{k,out}=\mathbf{m}_{j,out}$ (see Fig.~\ref{fig:split} and Fig.~\ref{fig:inout}).
\item \textbf{Case 2: no transition between $\mathbf{m}_i$ and $\mathbf{m}_j$ with $S(\mathbf{m}_i,\mathbf{m}_j) \geq t_s$.} 
$\mathbf{m}_i$, $\mathbf{m}_j$ and $\mathbf{m}_k$ are fused and added to the fusion set. The fused primitive is indexed as the minimum over the indices $i,j,k$.
\item \textbf{Case 3: no transition between $\mathbf{m}_i$ and $\mathbf{m}_j$ with $S(\mathbf{m}_i,\mathbf{m}_j) < t_s$.} This case happens when two motion primitives are matched to a portion of another primitive. In our experiments, we observed that such a case only arises when $t_s$ is too small and the matching is not valid. In this case the nodes and edges in this component are not updated.
\end{itemize}
\subsubsection{Connected components with three or more edges} In this case, the edge with the least similarity value is successively relaxed (removed) until the connected component has only two edges left. This relaxation does not affect performance as this situation rarely occurs. However, further investigation would benefit future work. 
 \begin{algorithm}[t]
 \caption{\small $M(n) = IncrementalLearning(M(n-1),M'(n))$} \label{alg:sila}
\textbf{Input:}  previous model $M(n-1)\{\mathbf{D},\mathbf{R}\}$ and incremental model $M'(n)\{\mathbf{D'},\mathbf{R'}\}$\\
\textbf{Output:}  updated model $M(n)\{\mathbf{D''},\mathbf{R''}\}$\\
$\hat{M}(n) \gets \{\hat{D}\{D:D'\}, \hat{R}\{R:R'\}\}, \emptyset \gets FuseSet$\\
$[\hat{D},\hat{R}] =Reindex(\hat{D},\hat{R})$\\ \label{line:initialize}

$G_{s} =SimilarityGraph(D,D')$\\ \label{line:similarity}
$CC =ConnectedComponents(G_{s})$\\
 \For {$k = 1 :|CC|$}{ \label{line:casesstart}
 $ee = |edges(CC_{k})| , [h,c] = nodesIdx(CC_{k})$\\
   \If{$ee==1$}{
          $ \mathbf{m}_{i,in} =\mathbf{m}_{i,out} = min( \mathbf{m}_{h,in},\mathbf{m}_{c,in}) , i \in \{h,c\}$ \label{lst:line:case1}
          \\ $FuseSet \gets FuseSet \cup  \mathbf{m}_{h} \cup \mathbf{m}_{c}$
    }
   \ElseIf{$ee==2$}{
         \For {$\mathbf{m}_i \in  nodes(CC_{k})$ \label{line:case2start}}    {
		    \If {$ Degree(\mathbf{m}_i) == 2$}{  $[w,q]=nbrsIdx(\mathbf{m}_i)$ *** neighbor index\\
		    	\If {$(\mathbf{m}_{w},\mathbf{m}_{q}) \in \mathbf{\hat{R}}$}{
		    		$\mathbf{m}_{i,in} \gets \mathbf{m}_{w,out}$\\ $\mathbf{m}_{i,out}\gets \mathbf{m}_{q,in} $
		    	}
		     	\ElseIf {$S(\mathbf{m}_{w},\mathbf{m}_{q}) \geq t_s$}{
		     	    $\forall j \in \{w,q,i\},  \mathbf{m}_{j,in} =\mathbf{m}_{j,out} =  min(\mathbf{m}_{w,in}, \mathbf{m}_{q,in},\mathbf{m}_{i,in})$\\
                    $FuseSet \gets FuseSet \ \cup \mathbf{m}_{i} \cup \mathbf{m}_{w} \ \cup  \mathbf{m}_{q}$
                }
            }
  	    }
    }
    
    \Else{
        $CC_{k} = Relaxedges(CC_{k},2)$\\
          Go to line \ref{line:case2start}
    }
  \label{line:casesend}
  }
  $\mathbf{D''}=FusePrimitives(FuseSet)$ \label{line:fusestart}\\
  $\mathbf{R''} = Reindex\&FuseEdges(\mathbf{\hat{R}},\mathbf{D''})$ \label{line:fuseend}
\\\Return $M(n)\{\mathbf{D''},\mathbf{R''}\}$
\end{algorithm}
\begin{algorithm}[t]
 \caption{\label{alg:fusefunc}\small $\mathbf{R''}=Reindex\&FuseEdges(\mathbf{\hat{R}},\mathbf{D''}\{\mathbf{m}_{k=1:N_1+N'_1}\})$} 
$\emptyset \gets \mathbf{R''}_{edges} ,\emptyset \gets \mathbf{R''}_{GP}, \emptyset \gets \mathbf{\tilde{R}}_{edges},  \emptyset \gets \mathbf{\tilde{R}}_{GP} $\\
\For {$(i,j) \in \mathbf{\hat{R}}_{edges}$} {
$\mathbf{\tilde{R}}_{edges} \gets \mathbf{\tilde{R}}_{edges} \cup (\mathbf{m}_{i,out}, \mathbf{m}_{j,in})$ }
    $[\mathbf{I},\mathbf{U}]= uniqueEdges(\mathbf{\tilde{R}}_{edges})$  *** $U$ is the unique edges\\
    \For {$i=1: |\mathbf{U}|$} {
$\mathbf{R''}_{edges} \gets \mathbf{R''}_{edges} \cup U_{i} $\\
$\mathbf{R''}_{GP} \gets \mathbf{R}_{GP} \cup FuseGP(\cup \mathbf{\hat{R}}_{j\in \mathbf{I}_{i},GP})$} *** $\mathbf{I}_i$ is the indexes of similar edges in original $\mathbf{\tilde{R}}_{edges}$, which is the same as $\mathbf{\hat{R}}_{edges}$
\\\Return $\mathbf{R''}$ 
\end{algorithm}
\subsection{Model Fusion}
\subsubsection{Fusion of nodes (motion primitives)}
\begin{figure}[h]
\centering
\includegraphics[width= \textwidth]{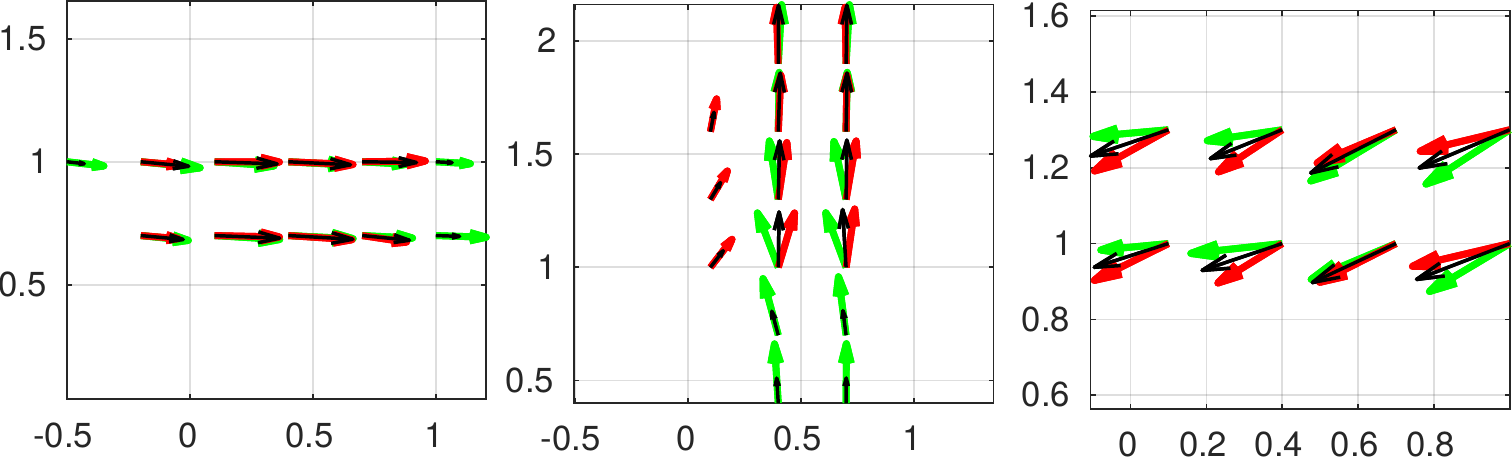}
\caption{\label{fig:similarity}  matched (similar) motion primitives learned from model $M(n-1)$ (green) and model $M'(n)$ (red). They are fused by considering the cell-wise average of each vector (black). Here, similarity threshold is $t_s =0.6$.}
\end{figure}

Motion primitives are fused according to the fusion set. The average of $\mathbf{m}_i$ and $\mathbf{m'}_j$ gives the fused motion primitive $\mathbf{m''}_i$ (Fig.~\ref{fig:similarity}).  Fused nodes and other nodes with no fusion are stored in $\mathbf{D''}$.
\subsubsection{Fusion of edges (transitions)}
To fuse transitions, each edge $(i,j)$ is first mapped to $(\mathbf{m}_{out,i}, \mathbf{m}_{in,j})$. This re-indexing can create similar edges and it happens when two nodes corresponding the indices are similar (Fig.~\ref{fig:graphunion}. Similar edges (including self-transitions) are then fused to get a `unique' list of edges and build $\mathbf{R''}_{edges}$ (Alg.~\ref{alg:fusefunc}).
To build $\mathbf{R''}_{GP}$, recall each transition is modeled as a pair of two dimensional GP flow fields ($GP_x$,$GP_y$), where each GP is represented by a sparse number of \emph{pseudo inputs}. Let the pseudo inputs and hyperparameters for matched transitions in $M(n-1)$ be denoted by $gp_1$ and the data associated with these transitions in $M'(n)$ be denoted by $d_2$. To fuse GPs associated with similar transitions in $M(n-1)$ and $M'(n)$, Ref.~\cite{suk2012incremental} is used to update $gp_1$ incrementally using $d_2$. Updated transitions are stored in $\mathbf{R''}:\{\mathbf{R''}_{edges},\mathbf{R''}_{GP}\}$

%% file: result.tex
\section{EXPERIMENTAL RESULT}
\subsection{Dataset Description}
\begin{figure*}
\includegraphics[width=0.165\textwidth]{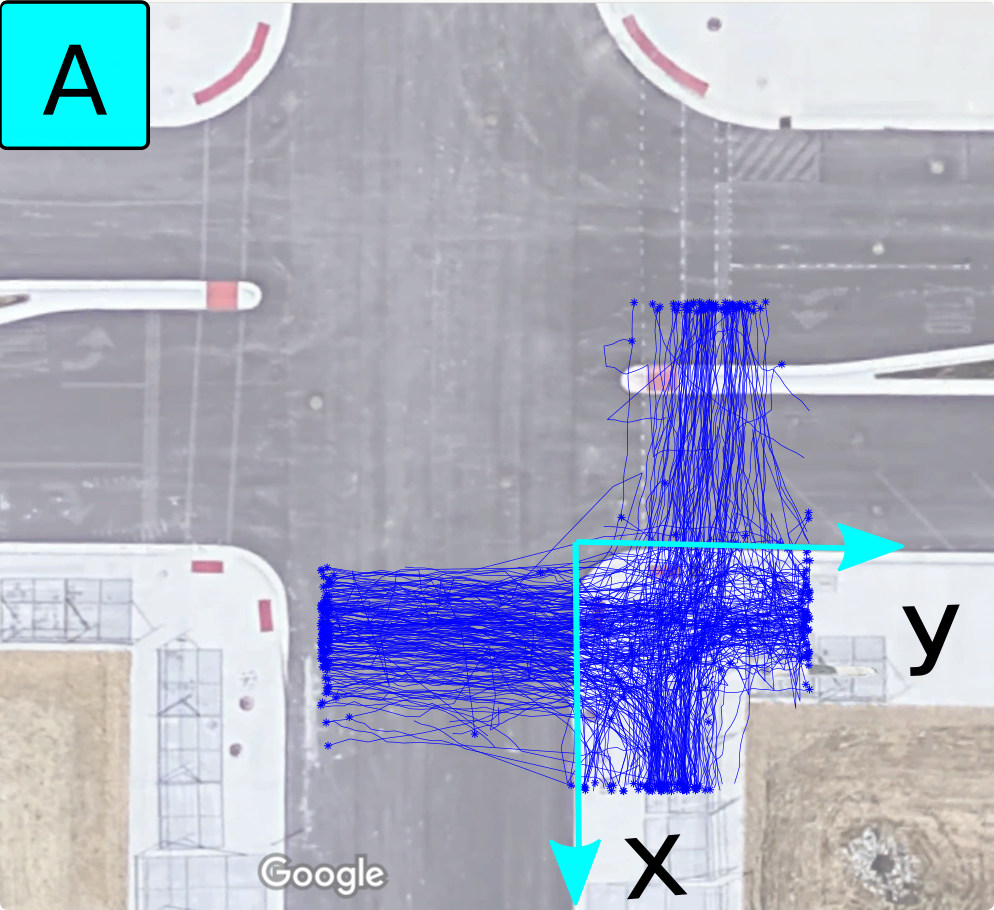}
\includegraphics[width=0.17\textwidth]{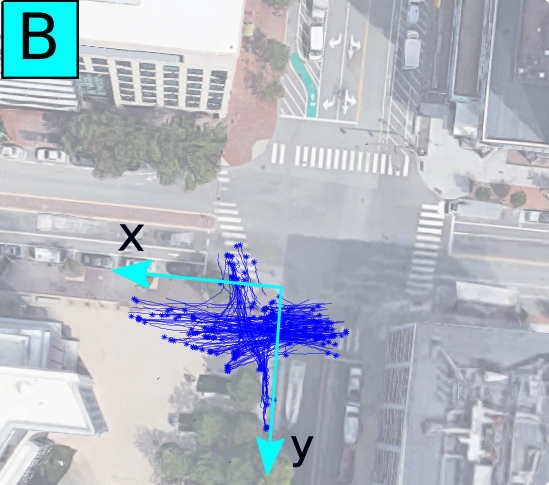}
\includegraphics[width=0.165\textwidth]{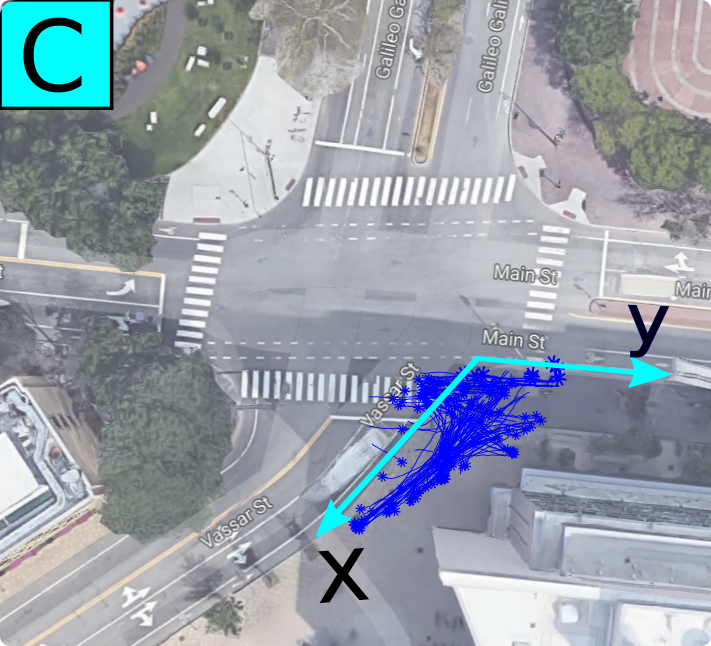}
\includegraphics[width=0.18\textwidth]{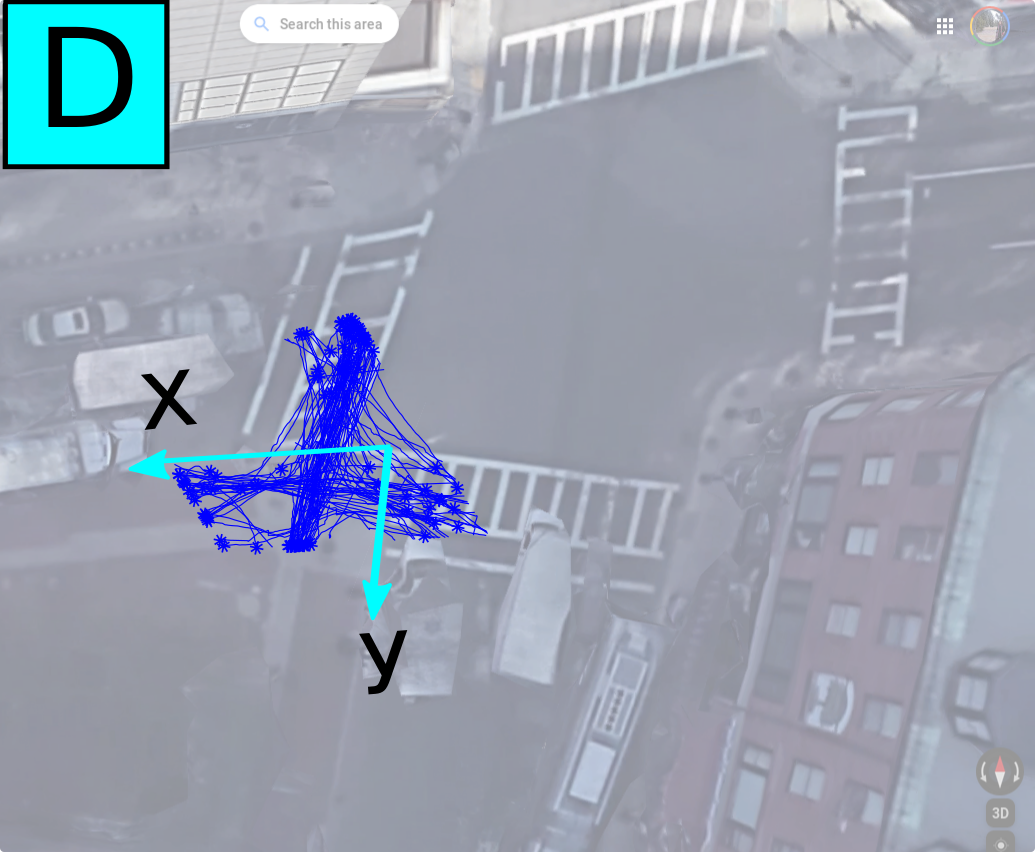}
\includegraphics[width=0.192\textwidth]{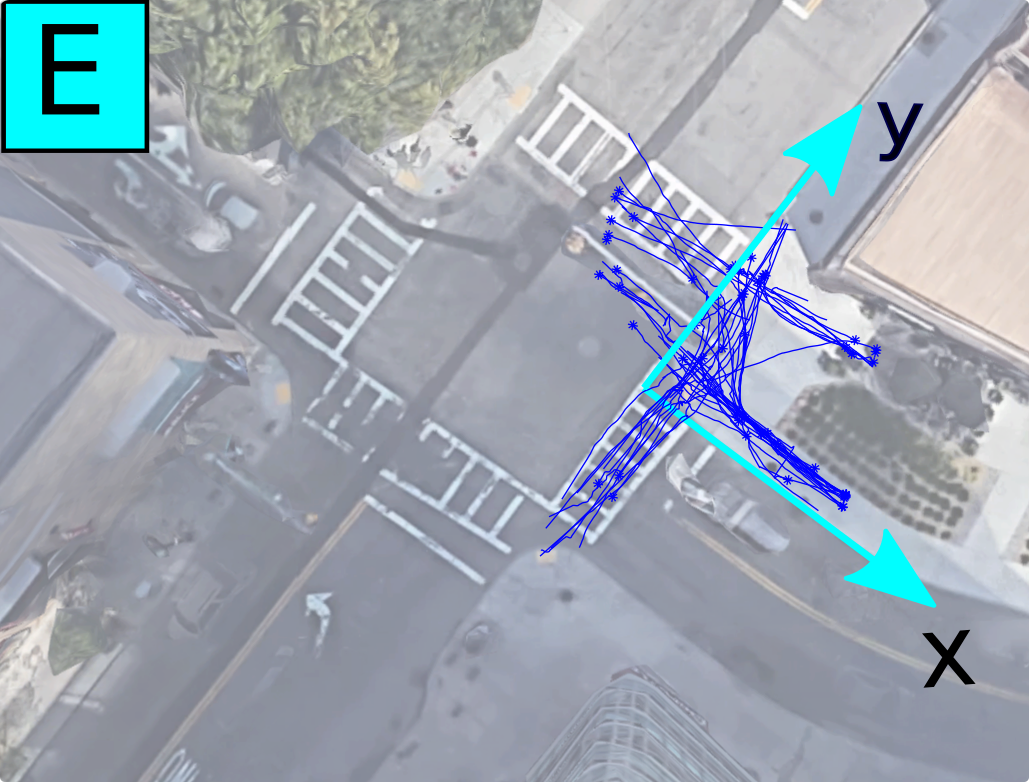}

\caption{\label{fig:intersections} Pedestrian trajectory data is collected in 5 intersections denoted by letters $A-E$, with different geometries (right, closed and open angle corners). Intersections $B-E$ are located in Cambridge/Boston areas, and intersection $A$ is in a mock city of  Mcity, The intersections and trajectories are shown in bird's eye view Google map, however the trajectory data is collected locally at the ground of each intersection, except data in  $A$ which is from GPS data. The intersection axis (cyan arrows) with origin at its corner is used for normalizing the trajectories described in section~\ref{sec:normalize}} 
\end{figure*}
\begin{figure*}
\subfigure[]{\label{fig:mhd_bigdata}
\includegraphics[width=0.24\textwidth]{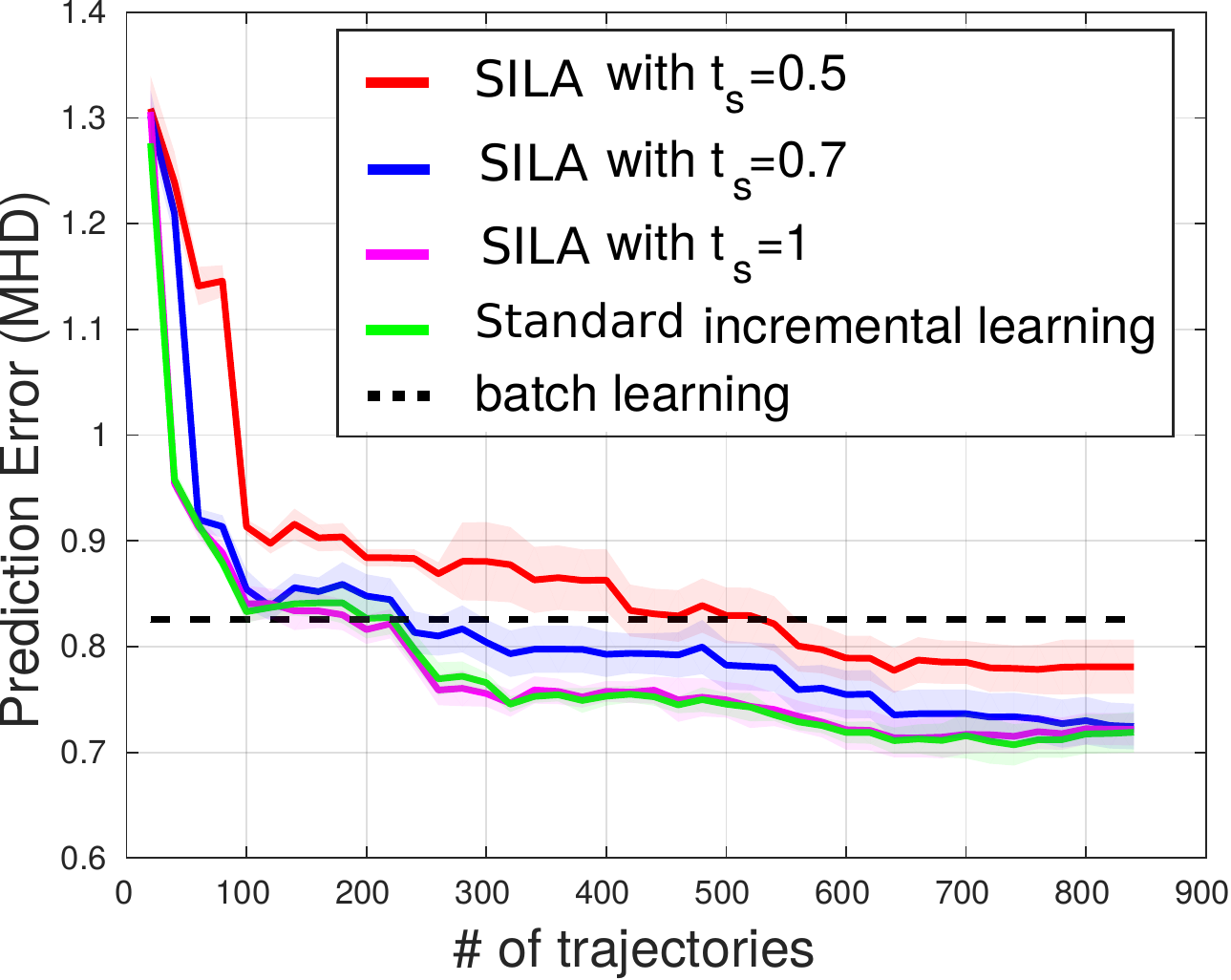}}
\subfigure[]{\label{fig:sizemhd_bigdata}
\includegraphics[width=0.24\textwidth]{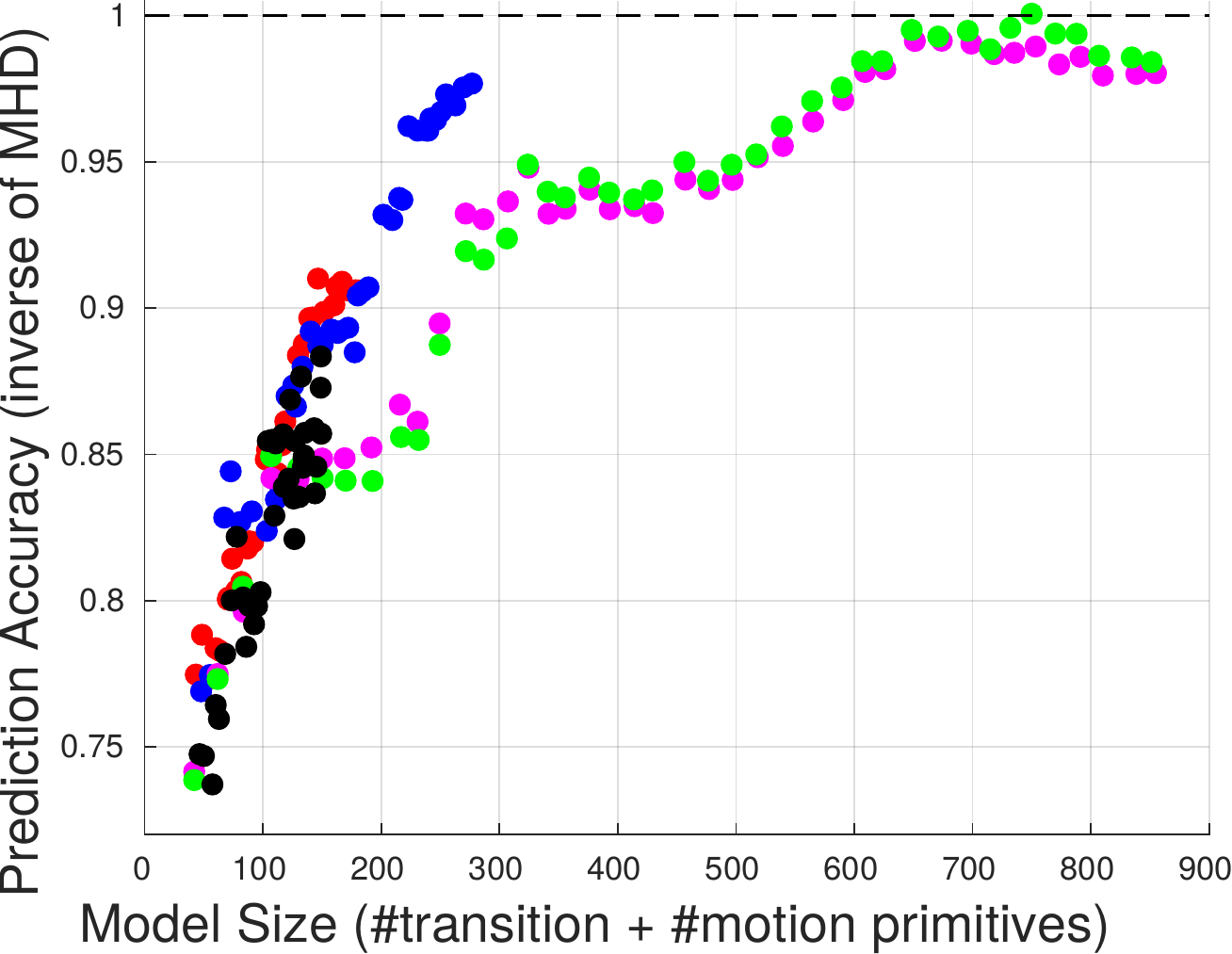}}
\subfigure[]{\label{fig:time_bigdata}
\includegraphics[width=0.24\textwidth]{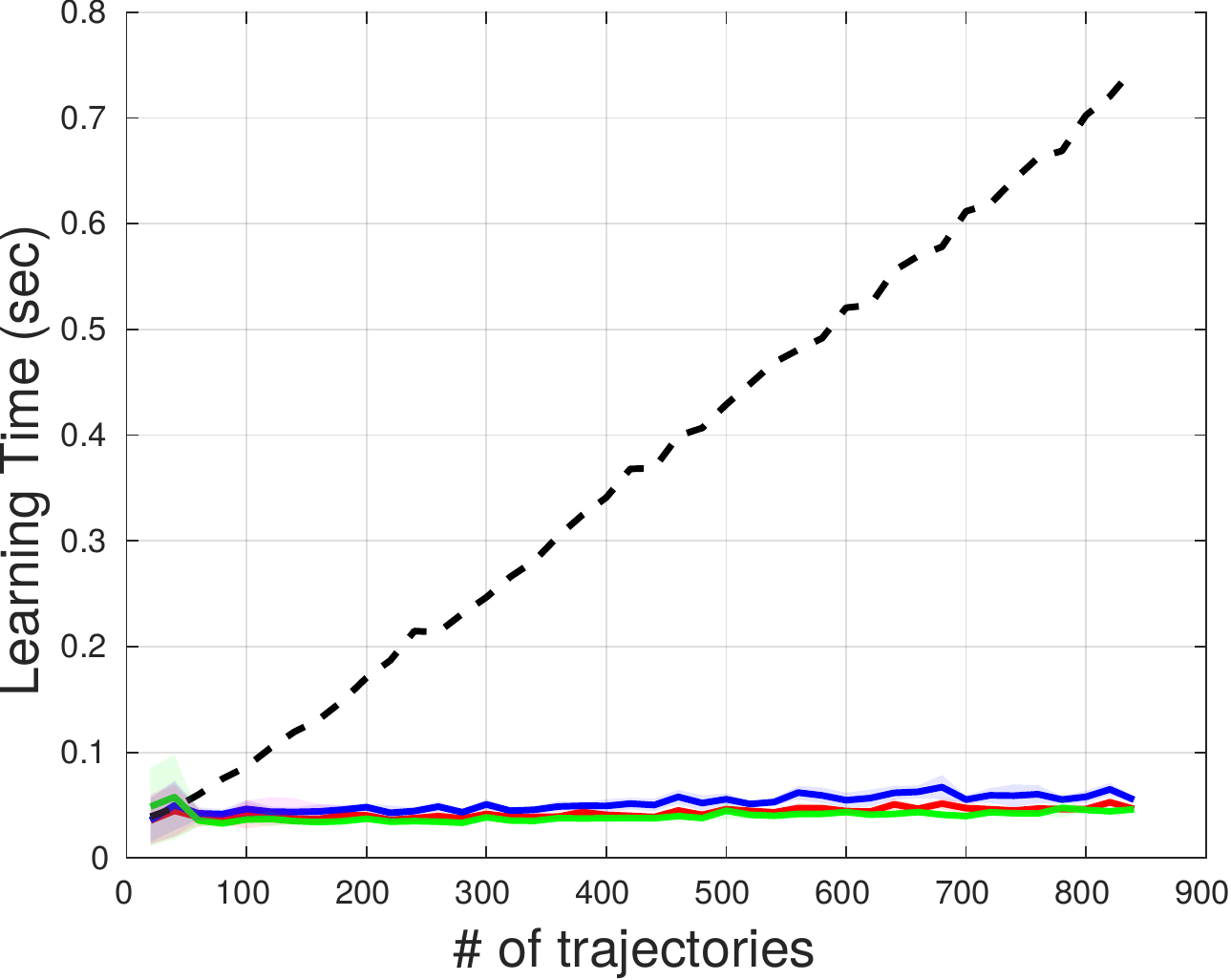}}
\subfigure[]{\label{fig:mhd_intersectiondata}\includegraphics[width=0.24\textwidth]{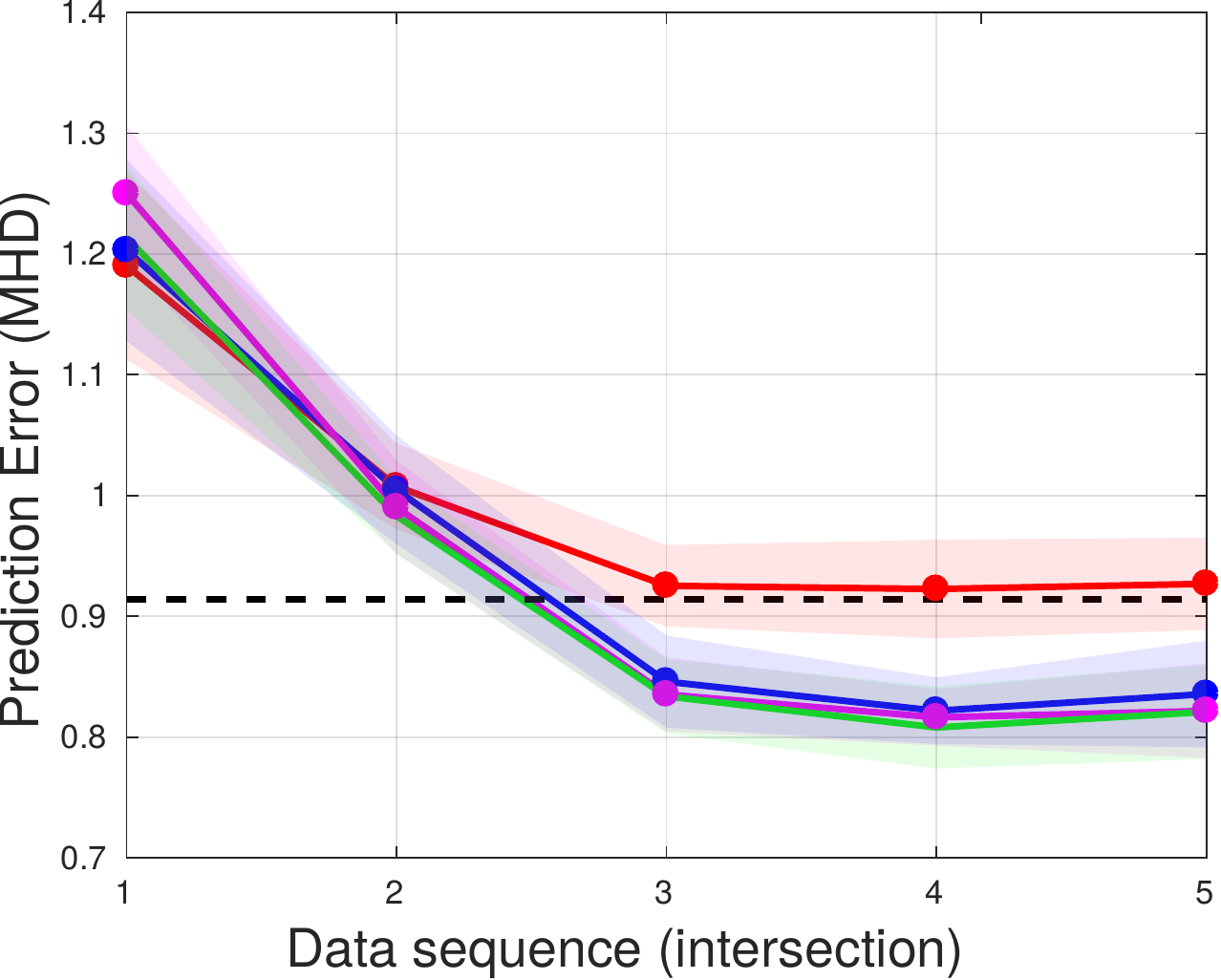}}
\caption{\label{fig:experiment} The performance comparison of SILA with standard incremental learning when the data is incrementally available over the time in one intersection in (a-c) and across 5 intersections (d), the result also compared with batch learning (dashed line), the results of average and standard deviation for 12 trials are compared for different algorithm (a-d): (a) prediction error  (MHD) (b) Prediction accuracy (normalized of inverse of MHD) vs model size (top left is better) (c) Learning time (sec)(d) prediction error (MHD) when data is available incrementally by exploring different  intersections shown in Fig.~\ref{fig:intersections}}
\end{figure*}
\begin{figure}
\includegraphics[trim=0 100 0 50,clip, width=0.4\textwidth] {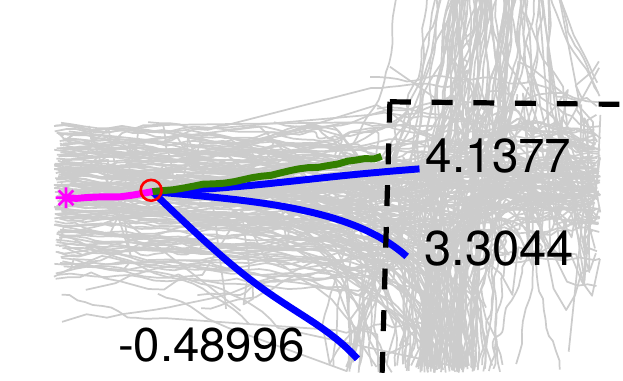}
\caption{\label{fig:metric}Example of prediction at an intersection (sidewalk is dashed line). Trajectory observed for 3.2s (magenta) and  predicted for 5s, which may contain multiple predicted trajectories with different probabilities (blue lines, numbers give negative log likelihood). To measure the prediction error respect to the ground truth (green line), the weighted average of prediction error (MHD) is computed, with weights corresponding to the likelihood of each prediction. Distribution of training data shown by grey lines.} 
\end{figure}
SILA and baselines are trained and evaluated on real pedestrian trajectory datasets collected in five intersections with different geometries and different neighborhood which creates various pedestrian behaviors (see Fig.~\ref{fig:intersections}).  
The datasets are also different in terms of measurement noise and quality. The trajectories in intersections $B$ and $C$ were collected using a GEM vehicle, equipped with 2D Lidar and cameras~\cite{miller2017predictive}, and curbside boundaries is extracted from the map created by Lidar. Trajectories in Intersections $E$ and $D$ were collected by a stationary tripod with a 3D Lidar. An online human classification technique, as proposed in~\cite{yan2017online}, was used to detect and track pedestrians. Curbside boundary was extracted by tracking a volunteer that walked the curbside boundary. 
\subsection{Baselines for Comparison}
In all experiments, SILA is run with the similarity thresholds $t_s= \in\{0.5,0.7,1.0\}$ and compared with the following two different baselines. 
\begin{itemize}
\item {\textbf{Baseline 1}-- Batch Learning.}
Batch Learning assumes the entire data is available and model is learned offline.
\item {\textbf{Baseline 2}-- Standard Incremental Learning.}
This algorithm is similar to SILA, there is no access to entire data because of memory constraints or the nature of data which is incrementally available only. To update the pre-trained model, the newly learned motion primitives and transitions from $M'(n)$ are simply added to the pre-trained model $M(n-1)$, without any fusion. The model size thus grows as $|M(n)|=\sum_{i=1}^n |M(i)|$.
\end{itemize}
\subsection{Evaluation Metrics:}
In all experiments, the future trajectory is predicted for the time horizon of 5 seconds after the trajectory is observed for 3.2 seconds. Three metrics are used for the evaluation: (1) prediction error compared to the ground truth using modified Hausdorff distance (MHD) \cite{dubuisson1994modified}, similar to \cite{2018arXiv180400495S} we use the weighted average of predicted trajectories. The weight is corresponding to likelihood of each prediction (Fig.~\ref{fig:metric}). (2) Model size:total number of motion primitives and number of transitions (3) Learning time.
\subsection{Experiment 1: Incremental Learning from data collected at one intersection}
We collected 989 trajectories in an intersection next to MIT campus, at different times (see Fig.~\ref{fig:intersections}, intersection B). The test size of 149 trajectories and 840 trajectories are split into 42 batches of 20 trajectories. The batches of data are given to the learning process in $42$ episodes. 
Fig.~\ref{fig:experiment} (a-c) shows the result of experiment 1 which run for 12 trials. The order of batch data is shuffled in each trial and the average of performance (prediction error) and model size growth are recorded for different baselines. 
 
All baselines show that, as the learning progresses, the prediction error decreases with increased model size. OF these,  baseline 2 approach has the largest growing rate of $1.01$ which (one model size growth per one trajectory) which is expected as the learned motion primitives and their transitions are added up without merging in each training episode.
 The model size in SILA with $t_s=0.7$ on the other hand grows with the rate of $0.31$.
As expected, SILA in the extreme case of $t_s=1$ (two motion primitives are matched only if they are the identical, which rarely happens and hence none of the motion primitives from two models are fused) is very similar to the standard incremental learning.
By decreasing the threshold in SILA, the model size growth rate is reduced as the number of matched motion primitive increases and more motion primitives are fused. However, lowering the threshold could encourage motion primitives with different behavior to be fused, leading to an increase in the prediction error. Fig.~\ref{fig:sizemhd_bigdata} shows the relation of prediction accuracy (normalized inverse of error) and the model size. The ideal approach would have the maximum accuracy even when the model size is still small (\emph{i.e.}, top left corner). This result indicates that SILA ($t_s=0.7$, blue in Fig.~\ref{fig:sizemhd_bigdata}) has the best combined performance. 

Another benefit of incremental learning compared to batch incremental learning; where every time the model is reset and trained from the accumulated data) is the learning time Fig.~\ref{fig:time_bigdata} shows the learning time in batch incremental learning increases linearly as the learning progresses over the time, but in incremental methods (standard and SILA), learning time is essentially a constant that is proportional to the size of incremental data (20 trajectories).
\subsection{Experiment 2: Incremental Learning from data collected in different intersections}
The dataset used for this experiment has a total of $412$ trajectories 
Each of intersections $A$ to $D$ contributed $79$ train and $13$ test trajectories. Intersection $E$ has $56$ trajectories only and contributed $48$ train and $8$ test trajectories.

Fig.~\ref{fig:mhd_intersectiondata} shows results from a model that incrementally learns from data in 5 intersections. Note the improvement in prediction error over time. Similar to Experiment 1, decreasing $t_s$ leads to a slower rate of model growth. For case of $t_{s} =0.7$, model growth rate is $1.02$. The model size growth ratio is higher than in Experiment 1 due to the greater diversity in motion behaviors across different intersections. Model growth rate of baseline 2 is $1.42$ (worst) as it is directly correlated to the size of the accumulated model.

 Both experiments standard incremental learning (baseline 2) and SILA outperform the batch learning method in terms of prediction error. Batch learning on a large dataset can result in learning motion primitives, less accurately than incremental learning which increases the prediction error.

%% file: conclusion.tex
\section{CONCLUSION}
This work proposed Similarity-based Incremental Learning Algorithm (SILA) for prediction of pedestrians trajectory when the data is incrementally available in one or across multiple intersections. A new model is learned from new data and the similarity between motion primitives from new and pre-trained models is computed for fusion of two models. The result confirms that SILA is able to incrementally learn and improve the prediction accuracy over the time and outperform batch (off-line) learning in predicting the pedestrian trajectories in intersections, while the learning time is constantly small, which enables SILA for online learning applications. While prediction accuracy of SILA is comparable to standard incremental learning, where the newly learned model are simply accumulated, the model size growth ratio is up to $3$ times slower than standard approach.

SILA can be considered as a meta learning approach for learning new motion behaviors. The extension of SILA for multi-agent learning is left for further study.